%% file: neurips_2026.tex
\documentclass{article}

\usepackage[preprint]{neurips_2026}
\usepackage{natbib}
\usepackage[T1]{fontenc}    
\usepackage{hyperref}       
\usepackage{url}            
\usepackage{booktabs}       
\usepackage{amsfonts}       
\usepackage{nicefrac}       
\usepackage{microtype}      
\usepackage{xcolor}         
\usepackage{pznotations}
\usepackage{graphicx}
\usepackage{placeins}
\usepackage{listings}
\usepackage{caption}
\definecolor{codebg}{RGB}{248,248,244}
\definecolor{codecomment}{RGB}{0,128,0}
\definecolor{codekw}{RGB}{128,0,128}

\lstdefinestyle{appendixcode}{
	basicstyle=\ttfamily\scriptsize,
	backgroundcolor=\color{codebg},
	frame=single,
	rulecolor=\color{black},
	breaklines=true,
	breakatwhitespace=false,
	columns=fullflexible,
	keepspaces=true,
	showstringspaces=false,
	captionpos=b,
	xleftmargin=0.02\linewidth,
	xrightmargin=0.02\linewidth
}

\lstdefinestyle{pythoncode}{
	style=appendixcode,
	language=Python,
	keywordstyle=\color{codekw}\bfseries,
	commentstyle=\color{codecomment},
	stringstyle=\color{blue!60!black}
}

\lstdefinelanguage{yaml}{
	keywords={true,false,null,y,n,yes,no,on,off},
	keywordstyle=\color{codekw}\bfseries,
	sensitive=false,
	comment=[l]{\#},
	morecomment=[l]{\#},
	commentstyle=\color{codecomment},
	morestring=[b]',
	morestring=[b]"
}

\lstdefinestyle{yamlcode}{
	style=appendixcode,
	language=yaml
}

\title{MILP-Evo: Closed-Loop Fully Automatic Design of MILP Solvers}

\author{%
    Jinbiao Nie$^{1,2}$ \quad
    Kewei Feng$^{3,2}$ \quad
    Xiaoyuan Zhang$^{2}$\thanks{Xiaoyuan Zhang is the corresponding author.}\quad
    Shan Yin$^{1}$ \quad
    Zizhuo Wang$^{4}$ \quad
    Bin Dong$^{2,5}$ \\[0.6em]
    $^{1}$BUPT \quad
    $^{2}$BZA \quad
    $^{3}$BIT \quad
    $^{4}$CUHK-Shenzhen \quad
    $^{5}$PKU \\
}

\begin{document}

\maketitle

\begin{abstract}
    Machine learning methods have shown that data-driven policies can accelerate mixed-integer linear programming (MILP) solvers, but many such approaches remain difficult to inspect, adapt, and deploy because the learned policy is represented as an external predictor or other opaque model. By contrast, explicit solver logic is easier to understand and integrate, but is usually hand-designed rather than learned from solver feedback. We study whether the automatic design of MILP solver logic can instead be cast as LLM-guided closed-loop search over executable white-box components evaluated directly by end-to-end solver behavior. To this end, we propose a closed-loop program evolution framework for MILP solver auto-design, implemented through PySCIPOpt, and instantiate it on the joint design of a cut selector and a branching rule. Candidate programs are iteratively generated, loaded into SCIP, and evaluated by direct execution on MILP instances, with the resulting feedback guiding performance-based selection, targeted repair, diagnostic reflection, and diversity-aware population maintenance. The method outputs explicit solver components that can be inspected, modified, and deployed within standard solver workflows. Across four benchmark families, we find that LLM-guided program evolution can discover competitive domain-specialized policies in several settings.
\end{abstract}

\section{Introduction}
	Combinatorial optimization (CO) is a foundational and challenging area of mathematical optimization, with applications ranging from production planning and distribution scheduling to chip design~\citep{liu2008tsp,chen2010integrated,ma2019accelerating}. Many exact CO tasks are naturally modeled as mixed-integer linear programs (MILPs), a standard language for industrial and benchmark optimization instances~\citep{bengio2021machine,gleixner2021miplib}. Because broad classes of MILPs are NP-hard, the practical value of exact optimization depends heavily on the efficiency of modern solvers. Such solvers build on branch-and-bound (B\&B) and branch-and-cut, combining tree search with cutting planes, presolve, primal heuristics, and other internal mechanisms to obtain certifiably optimal solutions~\citep{achterberg2009scip}.
	
	Modern MILP solvers have historically relied on human-designed policies for these decisions. Such policies are widely deployed because they are compact, efficient, reliable, and naturally integrated into solver workflows. For example, branching and cut selection rules are often implemented as hard-coded scoring functions or carefully engineered procedures that reflect decades of optimization expertise. However, designing these policies requires substantial expert knowledge and manual tuning, and generic rules may miss distribution-specific structure in recurring problem families~\citep{bengio2021machine,lodi2017learning,kuang2024rethinking}. This tension has motivated a growing body of learning-based work that replaces or augments individual solver decisions with data-driven policies, especially for branching~\citep{gasse2019exact,gupta2020hybrid,lin2024cambranch}, cut management~\citep{huang2022learning,paulus2022learning,turner2023adaptive,wang2023learning,li2023learning,puigdemont2024learning}, and other modules such as diving heuristics and presolving~\citep{paulus2023learning,liu2024l2p}.
	
	Although learned policies can improve solving efficiency on problem distributions with shared structure, many of them are implemented as external neural or graph-neural predictors. This introduces a different set of challenges: the learned decision logic can be difficult to interpret, may require nontrivial training and inference infrastructure, and is not always easy to deploy as a native part of a production solver. Recent work has begun to address this issue by learning more explicit and interpretable solver policies. In particular, symbolic-policy methods such as Symb4CO and GS4CO learn compact branching rules that are lightweight, interpretable, and closer in spirit to traditional solver heuristics~\citep{kuang2024rethinking,kuang2024towards}. These works suggest that automatic algorithm discovery can produce white-box solver policies rather than only black-box predictors. However, existing interpretable approaches still primarily focus on a single solver module, most notably branching, and often optimize a policy through imitation or proxy objectives rather than through direct end-to-end solver execution. Recent work has also begun to study coupled B\&B decisions such as node and variable selection jointly~\citep{du2025learning}, but the broader question remains: Can LLMs directly search over executable solver logic, using end-to-end solver feedback to optimize interacting MILP components rather than separate predictors or isolated learned rules?
	
	This paper studies this question through LLM-guided executable search over solver programs implemented via PySCIPOpt. We focus on branching and cut selection as a practical testbed for this setting. Beyond their individual importance, these two components interact naturally within B\&B-based MILP solving: branching shapes the evolving search tree and LP contexts in which cuts are generated and applied, while cut selection changes bound progress, LP relaxations, and the candidate states seen by later branching decisions. Both are also accessible through direct PySCIPOpt implementation, making them suitable for studying end-to-end code-level search under realistic solver constraints. Our framework uses an LLM to propose, mutate, repair, and reflect on candidate solver programs, but uses SCIP execution rather than LLM self-assessment to determine their quality. Each candidate is loaded into SCIP through PySCIPOpt \citep{maher2016pyscipopt,achterberg2009scip}, evaluated by end-to-end solving behavior, and constrained by solver-interface validity. The output is an explicit code artifact with concrete control logic, making the resulting solver behavior easier to inspect, debug, modify, and deploy as a native solver component. We study this framework on four learn2branch problem families: set cover, combinatorial auctions, facility location, and independent set.
	
	Our contributions are:
	(i) we formulate MILP solver-component auto-design as LLM-guided closed-loop search over executable PySCIPOpt callback components evaluated by end-to-end SCIP execution;
	(ii) we instantiate this formulation for the joint design of cut selection and branching rules, two interacting decisions in branch-and-cut search;
	(iii) we introduce an execution-guided evolutionary loop combining program proposal, targeted repair, diagnostic reflection, performance-based selection, and diversity-aware population maintenance under solver-interface constraints;
	(iv) we evaluate the discovered components on four learn2branch benchmark families---set cover, combinatorial auctions, facility location, and independent set---and show competitive domain-specialized performance in several settings.

\section{Related Work}

\subsection{MILP Solver Learning}
	Learning-augmented MILP solvers replace hand-designed components of branch-and-cut with policies trained from solver data. The most established line studies variable branching: GNN policies over the variable-constraint bipartite graph made it possible to imitate strong branching on recurring MILP distributions \citep{gasse2019exact}, while later methods reduce deployment cost or improve sample efficiency and transfer through hybrid inference and contrastive augmentation \citep{gupta2020hybrid,lin2024cambranch}. A parallel line studies cutting-plane decisions. Early learned cut selectors rank local candidate cuts, often with imitation or lookahead signals \citep{huang2022learning,paulus2022learning}; more recent work broadens the control surface to separator configuration, cut removal, hierarchical cut selection, and global cut selection across the branch-and-cut tree \citep{turner2023adaptive,wang2023learning,li2023learning,puigdemont2024learning,zeng2025beyond}. These methods show that solver decisions expose learnable structure, but the learned object is typically a predictor or policy attached to one decision interface.

	Recent work has also moved beyond isolated branching toward broader learned B\&B control. Symbolic-policy methods such as Symb4CO and GS4CO learn compact branching rules that are closer to traditional solver heuristics than opaque neural predictors \citep{kuang2024rethinking,kuang2024towards}. Other approaches learn coupled decisions, such as node and variable selection \citep{du2025learning}, or represent the entire B\&B tree to select search nodes with reinforcement learning \citep{zhang2025learning}. This progression is important for our setting because branching, node selection, and cut management interact through the solver trajectory rather than through independent one-step decisions. However, these works still primarily optimize learned decision models or symbolic formulas, whereas our goal is to search directly over executable callback code that jointly implements interacting solver modules.

	LLMs have recently entered MILP research through complementary routes. MILP-Evolve uses an LLM-based evolutionary process to generate diverse MILP problem classes, enabling foundation-model training for integrality-gap prediction, learning to branch, and instance-text alignment across problem families \citep{li2024towards}. LLM-LNS instead uses a dual-layer self-evolutionary LLM agent to design neighborhood-selection strategies for large-neighborhood search on large-scale MILPs \citep{ye2025large}. These studies suggest that LLMs can help expose optimization structure and synthesize useful solver heuristics. Our work differs in the target of the search: rather than generating training instances or controlling an LNS repair heuristic, we use LLM-guided execution feedback to evolve native PySCIPOpt branch-and-cut callbacks that run inside SCIP.

\subsection{Automatic Design}
	Evaluator-in-the-loop program search has emerged as a general mechanism for discovering executable heuristics and algorithms. FunSearch demonstrates that an LLM can propose small programs whose quality is determined by external execution rather than self-assessment \citep{romera2024mathematical}. Evolution of Heuristics and ReEvo adapt this idea to heuristic design by combining LLM mutation, selection, and reflective feedback over populations of candidate algorithms \citep{liu2024evolution,ye2024reevo}, while AlphaEvolve scales code evolution to scientific and algorithmic discovery tasks \citep{novikov2025alphaevolve}. The common pattern is to separate proposal from evaluation: the language model explores a program space, and an evaluator supplies the selection pressure.

	Our work is closest to this executable-discovery line, but the solver setting makes the search substantially more constrained than standalone heuristic design. A candidate program must be valid Python code, satisfy SCIP's callback contracts, return legal solver objects and result codes, and remain stable under repeated invocation inside a branch-and-cut solve. Moreover, the evaluated artifact is not a single scoring function for an external benchmark; it is a pair of interacting native solver modules whose effects are only visible through end-to-end SCIP execution. The contribution is therefore not generic LLM program search alone, but executable automatic design specialized to coupled MILP solver callbacks inside a standard optimization workflow.

\section{Method}
\label{sec:method}

\subsection{LLM-Guided MILP Solver Auto-Design}

We propose a framework for LLM-guided MILP solver auto-design: automatically constructing MILP solver logic by evolving native PySCIPOpt callback programs under end-to-end SCIP execution feedback. The key idea is to make the evolvable artifact an executable solver component rather than an external learned predictor. Candidate programs are proposed by an LLM, loaded into SCIP through the callback interface, evaluated by direct branch-and-cut execution, and selected according to measured solving behavior rather than LLM self-assessment.

This framing separates our setting from both learning-augmented MILP policies and generic LLM program search. Learned branching or cut-selection methods typically train a predictor for one solver decision interface, while our framework searches directly over executable solver components that run inside the solver. Generic evaluator-guided program search optimizes standalone code fragments \citep{romera2024mathematical,liu2024evolution,ye2024reevo,novikov2025alphaevolve}; in contrast, our candidates must satisfy solver-interface contracts and remain stable under repeated invocation during SCIP optimization. Thus the central difficulty is not only to generate code, but to generate admissible, executable, and useful solver logic.

\begin{figure}[t]
\centering
\includegraphics[width=\linewidth]{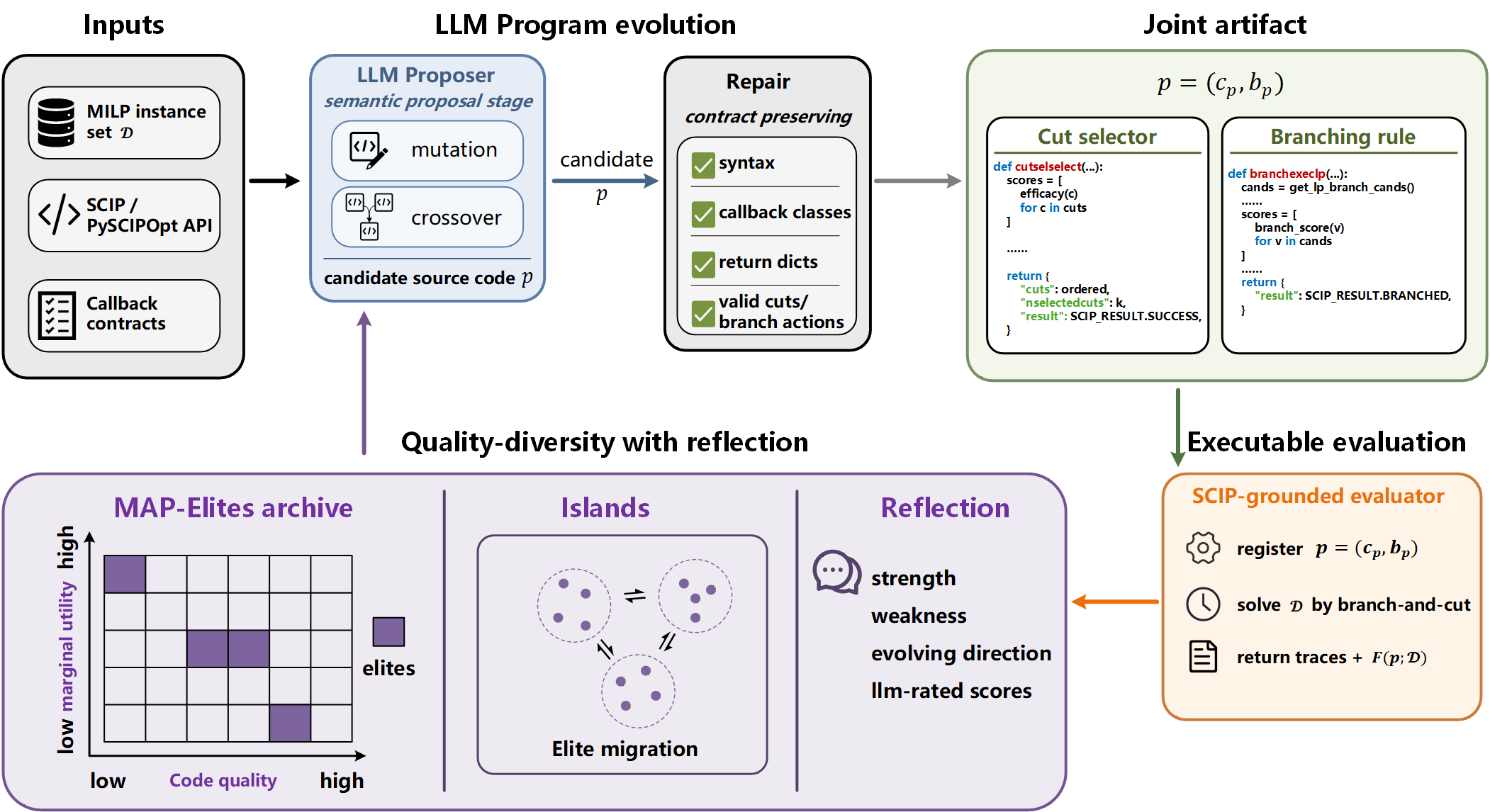}
\caption{Overview of the proposed framework. We adapt evaluator-guided code evolution to MILP solver auto-design by treating native PySCIPOpt callbacks as the evolvable artifact. The LLM proposes candidate source code, repair enforces callback contracts, the resulting joint artifact $p=(c_p,b_p)$ implements both cut selection and branching, and SCIP-grounded execution supplies traces and fitness for quality-diversity selection, island migration, and reflection.}
\label{fig:method-overview}
\end{figure}

Figure~\ref{fig:method-overview} summarizes the closed loop and its main artifacts. Starting from a MILP instance set, the SCIP/PySCIPOpt API, and callback contracts, the LLM proposes semantic edits to a joint callback program through mutation or crossover. Candidate code is then passed through contract-preserving repair and validation before it is admitted as a paired cut selector and branching rule. The repaired artifact is registered in SCIP, evaluated by branch-and-cut execution, and converted into traces and a scalar fitness $F(p;\mathcal{D})$; these signals update the MAP-Elites archive, island populations, and reflection summaries that condition later proposals.

\subsection{Joint Native Callback Search Space}

Each candidate is a joint native callback artifact: a user-defined Python program registered with SCIP and invoked by the solver at predefined control points of the optimization process. The callback artifact does not replace SCIP's main branch-and-cut loop; SCIP retains control of the solve and calls user logic only when a specific internal decision must be made.

In this work, each program is a joint artifact
\[
p = \big(c_p, b_p\big),
\]
where $c_p$ is a cut-selection callback and $b_p$ is a branching callback:
\begin{itemize}
\item The cut-selection callback is invoked after candidate cuts have been generated.  It scores, orders, and selects cuts from the candidate pool before they are added to the LP relaxation.
\item The branching callback is invoked when the solver needs to branch at an LP node.  It chooses a branching variable from the current fractional candidates.
\end{itemize}

The two callbacks are evolved as one source-code artifact and are loaded into the same SCIP model. This joint representation is essential because cut selection and branching are coupled through the solver trajectory. Cut selection changes the LP relaxation, bound progress, and candidate states later encountered by branching; branching changes the future nodes, LPs, and separation contexts in which cut selection is invoked. Searching over the pair $(c_p,b_p)$ therefore targets their combined effect on branch-and-bound rather than the isolated quality of either component.

This representation also keeps the learned policy explicit. The output of evolution is not a neural predictor wrapped around the solver, but readable solver logic implementing concrete decision rules. The program may use signals exposed by SCIP, such as cut efficacy, cut sparsity, row parallelism, objective coefficients, lock counts, pseudo-cost information, branching candidate statistics, and node-depth-dependent state. Evolution searches over how these solver signals are composed into cut-ranking, cut-filtering, variable-scoring, and depth-dependent control logic, while the callback interface constrains the program to remain deployable as a native solver component.

The search space is therefore not ordinary Python code. A generated source string first belongs to the unconstrained space $\widetilde{\mathcal{P}}$. It enters the effective search space only if it compiles, defines the required callback classes, satisfies the PySCIPOpt calling conventions, returns legal SCIP objects and result codes, selects cuts only from the candidate pool, and performs a valid branch action when the solver requires one. We define the feasible callback space operationally as
\[
\mathcal{P}
= \{p \in \widetilde{\mathcal{P}} : F(p;\mathcal{D}) < +\infty\},
\]
where infeasible programs are assigned infinite fitness and are excluded from elite selection. This view is important for solver auto-design: a candidate can express a useful algorithmic idea while still being inadmissible as a SCIP program because it violates a low-level callback contract. Invalid artifacts may fail through syntax errors, Python exceptions, malformed return dictionaries, illegal cut objects, degenerate branch behavior, or C-layer solver assertions.

The resulting optimization problem is
\[
p^\star \in \arg\min_{p \in \mathcal{P}} F(p;\mathcal{D}),
\]
where $F$ is induced by registering $p$ inside SCIP and running the resulting solver on MILP instances $\mathcal{D}$. Thus the language model supplies search proposals, while SCIP execution supplies both the feasibility test and the selection pressure.

\subsection{Contract-Preserving Evolution Loop}

Given a candidate program, the evaluator compiles the module, loads the callback artifact, checks that the required solver interfaces are present, and rejects programs that cannot be instantiated. For each MILP instance, it constructs a fresh SCIP model, registers the callbacks through PySCIPOpt, reads the instance, and invokes SCIP's optimization routine. The resulting trajectory provides process time, solve status, objective value when available, node count, solved-instance count, and structured error information.

We distinguish program-level callback validity from instance-level solve success. A program is callback-valid only if it satisfies the SCIP interface: the cut-selection callback must return cuts drawn from the original candidate pool, a legal selected-cut count, and a valid SCIP result code; the branching callback must perform a legal branch when LP branching candidates are available. Syntax errors, missing callback definitions, malformed return values, callback exceptions, invalid cut objects, and degenerate branching behavior are treated as program-level failures and receive $+\infty$ fitness.  Among callback-valid programs, an instance run is successful only when SCIP terminates with the required target status, which is optimality in our experiments.

Let $\mathcal{D}$ denote the evaluation batch. For a callback-valid program $p$ and instance $x$, let $T(p,x)$ be the measured process time of the SCIP run. The selection fitness is
\[
F(p;\mathcal{D}) =
\begin{cases}
\frac{1}{|\mathcal{D}|}\sum_{x\in\mathcal{D}}T(p,x),
& \text{if } p \text{ is callback-valid and succeeds on all } x\in\mathcal{D},\\
+\infty, & \text{otherwise}.
\end{cases}
\]
All other evaluation outcomes receive infinite fitness and cannot enter elite selection.  This objective makes the scalar selection signal depend on complete end-to-end solver behavior while keeping structured failure information available for repair and reflection. The evaluator also records solved count, status, node count, objective values, and error traces as diagnostics, while the scalar score remains grounded in SCIP execution.

The evolution loop separates proposing solver logic from making it executable. In the semantic proposal stage, the LLM is asked to improve the joint callback artifact: it may generate a new artifact, mutate a parent's scoring formulas and control schedules, or perform crossover by recombining useful motifs from two parents. These edits are represented as source code, but the intended object of search is the algorithmic logic: how to trade off cut efficacy and sparsity, how aggressively to remove nearly parallel cuts, how branching scores combine fractionality, locks, objective coefficients, and pseudo-costs, and how these decisions should change across the branch-and-cut tree.

When a candidate fails through a recoverable interface violation, we apply \emph{contract-preserving repair}. The repair operator receives the candidate code together with compiler errors, Python tracebacks, callback-wrapper diagnostics, solver status, and the relevant PySCIPOpt interface constraints. Its role is to project the proposed solver logic back into the feasible callback space, not to replace it with a new heuristic. In practice this means repairing imports, class definitions, method signatures, return dictionaries, SCIP result codes, branch actions, and illegal API uses while preserving the candidate's cut-ranking and branching-scoring logic as much as possible. The repaired program is then re-evaluated by the same SCIP-grounded objective before it can enter the archive.

Finally, the archive maintains both performance and diversity. Each evaluated individual stores source code, fitness, parentage, solver diagnostics, error traces, reflection summaries, and two language-model-rated descriptors: \emph{code quality} and \emph{marginal utility}. These descriptors affect the exploration geometry by assigning candidates to cells in a two-dimensional MAP-Elites grid; they do not decide which candidate is better inside a cell. Within each cell, replacement is determined solely by SCIP-grounded fitness $F$.

Within each island, each cell retains only a small number of finite-fitness elites. This prevents the search from collapsing immediately onto one visually similar family of programs, while still using solve time as the selection criterion inside each cell. Reflection converts solver feedback into strengths, weaknesses, and suggested evolving directions, which are fed back into later mutation and crossover prompts. Multiple islands evolve in parallel with periodic migration of high-quality individuals, providing a second source of diversity at the population level.

The final output is an explicit joint solver artifact that can be inspected, modified, and deployed through the same PySCIPOpt pathway used during evolution.

\section{Experiments}

\subsection{Experimental Setup}

\paragraph{Baselines.}
We compare against a set of representative branch-and-cut baselines covering classical solver rules, learned branching policies, hybrid methods, and symbolic branching policies. FSB is full strong branching, RPB is reliability pseudo-cost branching, GNN is the supervised graph neural branching policy \citep{gasse2019exact}, Hybrid combines learned and solver-side branching scores \citep{gupta2020hybrid}, and GS4CO is a white-box symbolic branching baseline \citep{kuang2024rethinking,kuang2024towards}. We also include default SCIP as a solver-default baseline. MILP-Evo is evaluated as the exported PySCIPOpt artifact produced by execution-guided search, jointly implementing a cut selector and a branching rule. The primary comparison set consists of FSB, RPB, GNN, Hybrid, GS4CO, SCIP, and MILP-Evo; GNN-GPU and COPT~\citep{copt} are included only as external references and are excluded from optimal-run win counts and boldface comparisons.

\paragraph{Benchmarks.}
We evaluate on four learn2branch MILP families: set cover, combinatorial auctions, facility location, and independent set \citep{gasse2019exact}. Each family contains Easy, Medium, and Hard transfer scales.

\paragraph{Training and evaluation.}
For each problem family, MILP-Evo uses Gemini-3-flash-preview as the language-model backbone and performs 400 search iterations on Easy-scale instances to obtain a domain-specialized joint callback program. The exported PySCIPOpt policy is then evaluated without further adaptation across the Easy, Medium, and Hard scales of the same family. All methods are measured by mean process time with a $3600$ second time limit, with unsolved runs charged the full time limit. We also report per-scale win counts over the primary comparison set and mean branch-and-bound node counts. MILP-Evo was run on machines equipped with Intel Xeon Platinum 8362 processors, using CPU execution only and no GPU acceleration. Its main software environment used Python 3.11.9, PySCIPOpt 6.1.0, SCIP 10.0.2, and SoPlex 8.0.2. Baseline implementations were run with the Python and package environments required by their original implementations.

\subsection{Comparative Evaluation and Search Behavior}

Table~\ref{tab:main-results} reports the main comparative results. Following prior branching-policy evaluations, each block gives mean process time, win count, and mean node count for one problem family across the three transfer scales.

\begingroup
\setlength{\abovecaptionskip}{0pt}
\setlength{\belowcaptionskip}{0.2em}
\begin{table}[!t]
	\centering
	\scriptsize
	\setlength{\tabcolsep}{3.2pt}
	\setlength{\aboverulesep}{0.12ex}
	\setlength{\belowrulesep}{0.24ex}
	\newcommand{\diffmidrule}{\cmidrule(lr){2-4}\cmidrule(lr){5-7}\cmidrule(lr){8-10}}
	\renewcommand{\arraystretch}{0.88}
	\caption{Comparative evaluation on four learn2branch families. Time is mean \texttt{proctime} in seconds; Wins and boldface are computed over primary comparison set only. GNN-GPU and COPT are external references.}
	\label{tab:main-results}
	\vspace{-0.45em}
	\resizebox{0.86\linewidth}{!}{%
		\begin{tabular}{lrrr@{\quad}rrr@{\quad}rrr}
			\toprule
			Indset: & \multicolumn{3}{c}{Easy} & \multicolumn{3}{c}{Medium} & \multicolumn{3}{c}{Hard} \\
			\midrule
			Model & Time(s) & Wins & Nodes & Time(s) & Wins & Nodes & Time(s) & Wins & Nodes \\
			\diffmidrule
			FSB & 87.00 & 1/50 & 30.2 & 3318.97 & 0/7 & 150.5 & 3600.06 & 0/0 & 53.9 \\
			\diffmidrule
			RPB & 9.49 & 0/50 & 46.9 & 194.74 & 0/50 & 7442.9 & 2587.99 & 0/15 & 75329.3 \\
			GNN & 5.80 & 0/50 & 62.2 & 1673.24 & 0/36 & 19872.9 & 3600.01 & 0/0 & 49010.6 \\
			Hybrid & 6.47 & 0/50 & 53.1 & 1502.27 & 0/30 & 48887.3 & 3600.01 & 0/0 & 42854.8 \\
			GS4CO & 7.29 & 0/50 & 54.1 & 147.26 & 0/50 & 2785.0 & 2057.08 & 0/30 & 67083.8 \\
			SCIP & 14.80 & 0/50 & 164.6 & 377.99 & 0/50 & 4023.9 & 3155.52 & 0/18 & 22613.1 \\
			MILP-Evo & \textbf{2.57} & \textbf{49/50} & 84.7 & \textbf{59.20} & \textbf{50/50} & 1051.9 & \textbf{806.17} & \textbf{50/50} & 17980.3 \\
			\diffmidrule
			GNN-GPU & 4.63 & -/50 & 62.2 & 858.07 & -/45 & 57818.8 & 3600.00 & -/0 & 107167.4 \\
			COPT & 2.62 & -/50 & 9.6 & 62.18 & -/50 & 707.7 & 1531.27 & -/43 & 10279.2 \\
			\midrule\midrule
			Cauctions: & \multicolumn{3}{c}{Easy} & \multicolumn{3}{c}{Medium} & \multicolumn{3}{c}{Hard} \\
			\midrule
			Model & Time(s) & Wins & Nodes & Time(s) & Wins & Nodes & Time(s) & Wins & Nodes \\
			\diffmidrule
			FSB & 10.80 & 0/50 & 25.5 & 286.99 & 0/50 & 253.1 & 3177.54 & 0/14 & 912.3 \\
			\diffmidrule
			RPB & 3.70 & 0/50 & 14.8 & 21.69 & 1/50 & 1169.0 & \textbf{271.44} & \textbf{36/50} & 17918.5 \\
			GNN & 3.13 & 0/50 & 105.8 & 30.05 & 0/50 & 1047.9 & 662.39 & 0/50 & 17571.3 \\
			Hybrid & 2.07 & 0/50 & 111.2 & 21.26 & 0/50 & 1210.7 & 718.47 & 0/50 & 30404.5 \\
			GS4CO & 3.07 & 0/50 & 114.3 & 42.55 & 0/50 & 1353.1 & 1248.98 & 0/45 & 24867.6 \\
			SCIP & 3.57 & 0/50 & 30.1 & 44.96 & 0/50 & 1169.9 & 307.55 & 3/50 & 15560.5 \\
			MILP-Evo & \textbf{0.81} & \textbf{50/50} & 196.1 & \textbf{11.79} & \textbf{49/50} & 1536.0 & 381.90 & 11/50 & 34534.7 \\
			\diffmidrule
			GNN-GPU & 1.60 & -/50 & 105.8 & 10.36 & -/50 & 1047.9 & 191.11 & -/50 & 17571.7 \\
			COPT & 2.18 & -/50 & 126.8 & 14.11 & -/50 & 1542.2 & 161.34 & -/50 & 19687.7 \\
			\midrule\midrule
			Setcover: & \multicolumn{3}{c}{Easy} & \multicolumn{3}{c}{Medium} & \multicolumn{3}{c}{Hard} \\
			\midrule
			Model & Time(s) & Wins & Nodes & Time(s) & Wins & Nodes & Time(s) & Wins & Nodes \\
			\diffmidrule
			FSB & 29.07 & 0/50 & 45.7 & 1291.03 & 0/45 & 929.5 & 3600.03 & 0/0 & 658.3 \\
			\diffmidrule
			RPB & 11.20 & 0/50 & 129.6 & \textbf{162.69} & \textbf{29/50} & 10349.9 & 2401.92 & \textbf{19/20} & 118275.4 \\
			GNN & 11.32 & 3/50 & 172.7 & 487.64 & 0/50 & 5990.5 & 3500.75 & 0/5 & 21036.2 \\
			Hybrid & \textbf{8.26} & \textbf{29/50} & 201.9 & 171.41 & 15/50 & 7240.7 & \textbf{2359.13} & 0/10 & 63408.9 \\
			GS4CO & 13.89 & 0/50 & 249.1 & 609.94 & 0/45 & 10303.9 & 2937.46 & 0/5 & 30500.2 \\
			SCIP & 14.18 & 0/50 & 124.5 & 274.02 & 0/50 & 8361.4 & 2955.81 & 0/20 & 115143.9 \\
			MILP-Evo & 9.03 & 18/50 & 232.6 & 166.69 & 6/50 & 10109.2 & 2895.95 & 5/24 & 133883.2 \\
			\diffmidrule
			GNN-GPU & 5.31 & -/50 & 172.7 & 102.20 & -/50 & 5990.5 & 2492.68 & -/20 & 102635.8 \\
			COPT & 6.88 & -/50 & 297.4 & 99.64 & -/50 & 12207.1 & 2718.71 & -/25 & 175771.1 \\
			\midrule\midrule
			Facilities: & \multicolumn{3}{c}{Easy} & \multicolumn{3}{c}{Medium} & \multicolumn{3}{c}{Hard} \\
			\midrule
			Model & Time(s) & Wins & Nodes & Time(s) & Wins & Nodes & Time(s) & Wins & Nodes \\
			\diffmidrule
			FSB & 81.02 & 0/50 & 71.0 & 806.48 & 0/50 & 233.9 & 2108.21 & 0/41 & 109.6 \\
			\diffmidrule
			RPB & 36.03 & 16/50 & 90.4 & 305.37 & 6/50 & 588.6 & \textbf{607.57} & 15/50 & 260.1 \\
			GNN & 50.48 & 0/50 & 287.6 & 478.05 & 7/50 & 881.4 & 669.15 & 8/50 & 462.3 \\
			Hybrid & \textbf{35.09} & \textbf{17/50} & 253.2 & 325.92 & 14/50 & 895.3 & 631.09 & \textbf{17/50} & 456.6 \\
			GS4CO & 68.69 & 0/50 & 273.6 & 558.56 & 1/50 & 933.8 & 731.28 & 4/50 & 449.9 \\
			SCIP & 38.27 & 5/50 & 99.5 & 305.10 & 2/50 & 541.1 & 675.33 & 3/50 & 194.6 \\
			MILP-Evo & 35.66 & 12/50 & 235.5 & \textbf{266.21} & \textbf{20/50} & 926.8 & 761.88 & 3/50 & 429.1 \\
			\diffmidrule
			GNN-GPU & 33.44 & -/50 & 287.6 & 267.77 & -/50 & 881.4 & 546.13 & -/50 & 462.3 \\
			COPT & 8.20 & -/50 & 221.0 & 31.56 & -/50 & 572.0 & 58.06 & -/50 & 262.8 \\
			\bottomrule
		\end{tabular}%
	}
	\vspace{-0.6em}
\end{table}
\FloatBarrier
\vspace{-0.45em}
\endgroup

Overall, MILP-Evo is strongly competitive across the four benchmark families, with the clearest gains on independent set. On the remaining families, it also frequently reaches the best or near-best performance among the primary comparison methods.

These results suggest that joint executable search can discover domain-specific and effective cut-selection and branching rules, rather than only generic solver heuristics. At the same time, the gains are not uniform across all families and transfer scales, indicating that the effectiveness of the discovered rules remains somewhat uneven.

\paragraph{Search convergence}
Figure~\ref{fig:search-dynamics} summarizes both the optimization trace and the ancestry of the best independent-set artifact. The convergence curve is staircase-shaped rather than smooth: search spends long periods on valid but non-improving programs, then improves when a structural edit changes how cut filtering and branching interact.

\paragraph{Lineage analysis}
The lineage panel adds the mechanism behind this pattern. The final artifact does not descend through a monotone sequence of better programs. Exploratory recombination initially worsens immediate fitness while introducing reusable motifs such as sparsity-aware cut scoring and value-sensitive branching. Later descendants preserve these motifs and refine them into structure-aware branching, graph-connectivity signals, and calibrated cut filtering. The best program is therefore the result of non-monotone evolutionary reuse rather than a single isolated improvement.

\vspace{-0.25em}
\begin{center}
\refstepcounter{figure}\label{fig:search-dynamics}
\centering
\includegraphics[width=0.95\linewidth]{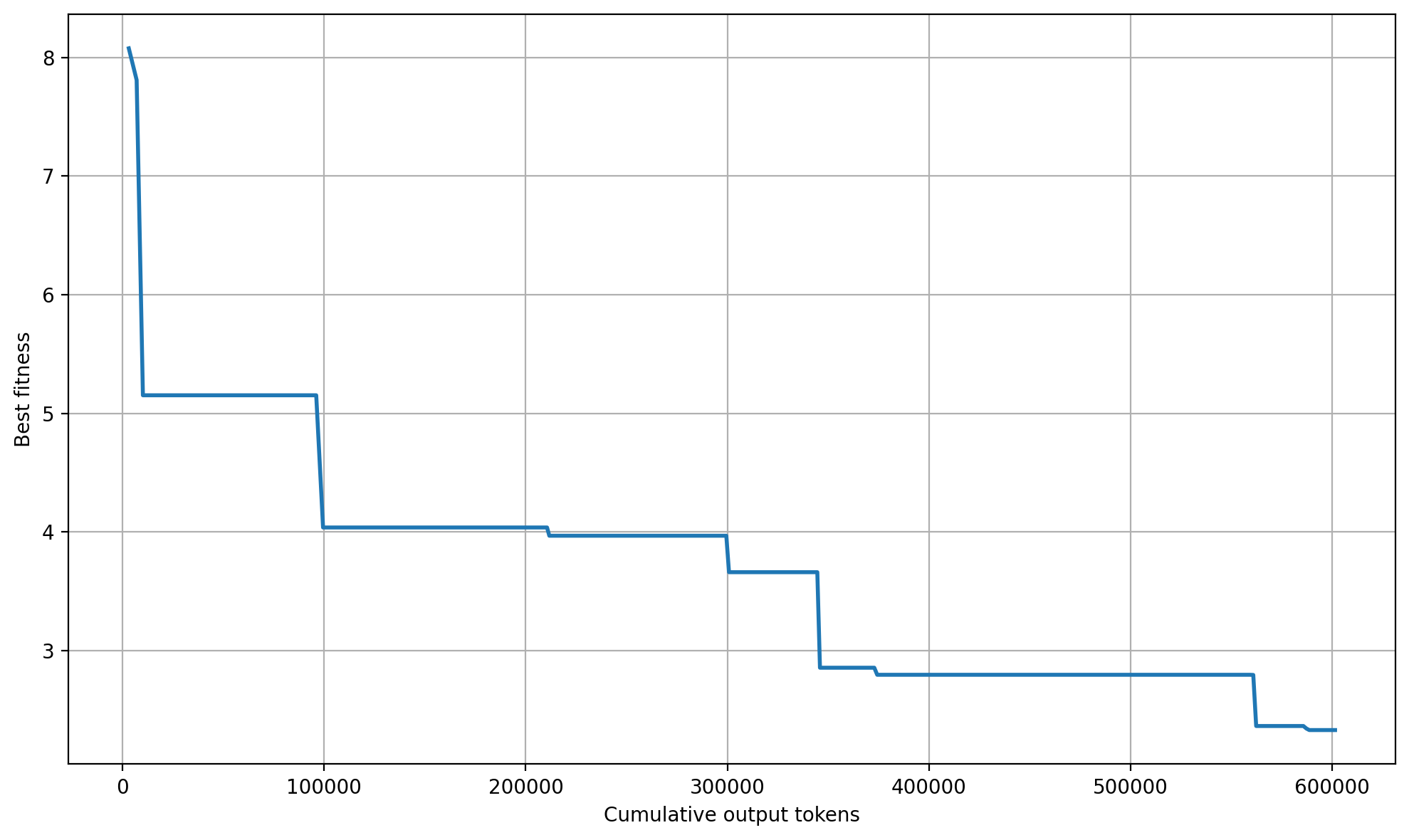}\\
{\footnotesize (a) Best fitness versus cumulative LLM output tokens.}\\[-0.2em]
\includegraphics[width=0.95\linewidth]{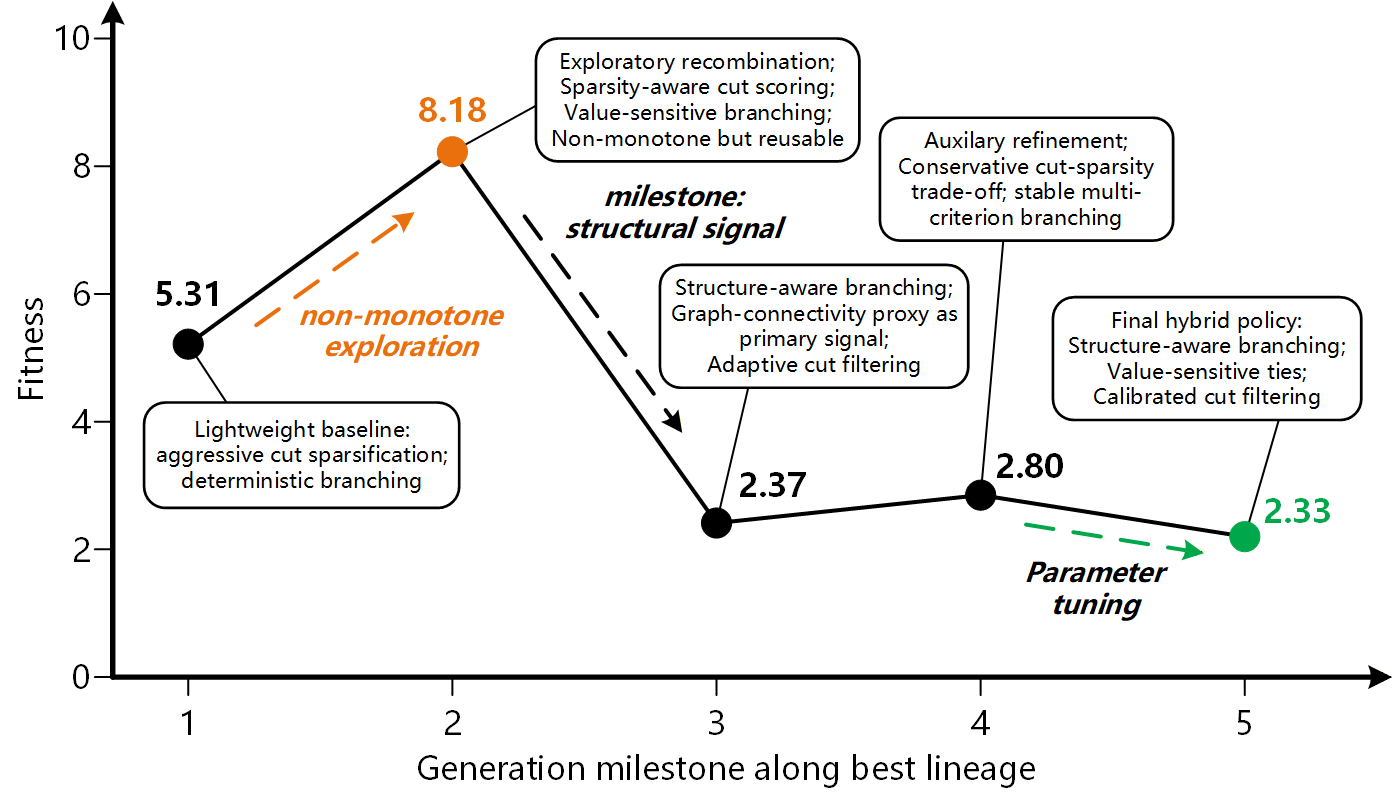}\\
{\footnotesize (b) Milestones along the best lineage.}\\[-0.2em]
{\footnotesize Figure~\thefigure: Search dynamics on independent set; lower fitness is better.}
\end{center}
\vspace{-0.85em}

\subsection{Ablation Study}

We use independent set as a representative case for the ablation study. The cut-only and branch-only variants isolate the two callback components. The separate-best variant combines the best independently found single-component artifacts after search, without co-evolving them as one program. NoRepair disables the contract-preserving repair step described in Section~\ref{sec:method}.

\begin{center}
\refstepcounter{table}\label{tab:ablation}
\scriptsize
\setlength{\tabcolsep}{3.2pt}
\renewcommand{\arraystretch}{0.9}
{\footnotesize Table~\thetable: Ablation on independent set. Time is mean \texttt{proctime} in seconds.}
\resizebox{0.88\linewidth}{!}{%
\begin{tabular}{lrrr@{\quad}rrr@{\quad}rrr}
\toprule
Variant & \multicolumn{3}{c}{Easy} & \multicolumn{3}{c}{Medium} & \multicolumn{3}{c}{Hard} \\
\midrule
 & Time(s) & Wins & Nodes & Time(s) & Wins & Nodes & Time(s) & Wins & Nodes \\
\cmidrule(lr){2-4}\cmidrule(lr){5-7}\cmidrule(lr){8-10}
Cut only & 6.33 & 4/50 & 45.1 & 179.96 & 0/50 & 2121.0 & 2522.53 & 0/28 & 22734.0 \\
Branch only & 4.85 & 1/50 & 137.1 & 166.36 & 0/50 & 4243.6 & 2452.77 & 0/26 & 46905.8 \\
Separate best & 3.95 & 7/50 & 132.9 & 124.48 & 4/50 & 4053.4 & 2299.23 & 0/29 & 52902.7 \\
SCIP & 14.80 & 0/50 & 164.6 & 377.99 & 0/50 & 4023.9 & 3155.52 & 0/18 & 22613.1 \\
NoRepair & 6.34 & 0/50 & 107.0 & 200.81 & 0/50 & 4016.6 & 2615.15 & 0/22 & 26773.5 \\
MILP-Evo & \textbf{2.57} & \textbf{38/50} & 84.7 & \textbf{59.20} & \textbf{46/50} & 1051.9 & \textbf{806.17} & \textbf{50/50} & 17980.3 \\
\bottomrule
\end{tabular}%
}
\end{center}
\vspace{-0.8em}
\FloatBarrier

The ablation supports the importance of joint executable search. MILP-Evo outperforms both single-component variants and the post-hoc separate-best combination on all three transfer scales, reducing Hard-scale time from $2299.23$ seconds for separate-best to $806.17$ seconds. Disabling repair is also damaging: NoRepair is slower than MILP-Evo on every scale and wins no instances, even on Easy. Runtime gains are again not explained only by smaller trees; the evolved policy changes the cost and quality of the explored tree as well as its size.

\paragraph{Repair effectiveness.}
Table~\ref{tab:repair} reports repair-oriented success statistics over the stored candidate records. Repair substantially improves the reliability of executable callback generation, increasing the aggregate success rate from $79.96\%$ to $95.64\%$ and reducing final failures to $60$ cases. The effect is especially visible on the families with lower pre-repair validity, where many otherwise discarded candidates can be projected back into the feasible callback space. Across all triggered repairs, roughly four out of five repaired candidates become final successful records. This suggests that repair is not only a convenience for handling syntax or interface mistakes, but also a stabilizing mechanism that lets the search retain useful solver logic proposed in imperfect executable form.

\vspace{-0.2em}
\begin{center}
\refstepcounter{table}\label{tab:repair}
\scriptsize
\setlength{\tabcolsep}{3.6pt}
\renewcommand{\arraystretch}{0.82}
{\footnotesize\parbox{0.80\linewidth}{\centering Table~\thetable: Repair success statistics. ``Without repair'' treats triggered repairs as pre-repair failures; ``with repair'' uses the final evaluation outcome.}\par}
\vspace{-0.35em}
\resizebox{0.80\linewidth}{!}{%
\begin{tabular}{lrrrr}
\toprule
Family & Without repair & With repair & Triggered repair & Repaired success \\
\midrule
indset & $88.07\%$ ($288/327$) & $98.47\%$ ($322/327$) & $38$ & $34/38$ ($89.47\%$) \\
facilities & $82.86\%$ ($290/350$) & $98.57\%$ ($345/350$) & $58$ & $55/58$ ($94.83\%$) \\
cauctions & $79.19\%$ ($312/394$) & $94.42\%$ ($372/394$) & $82$ & $60/82$ ($73.17\%$) \\
setcover & $68.95\%$ ($211/306$) & $90.85\%$ ($278/306$) & $94$ & $67/94$ ($71.28\%$) \\
\midrule
overall & $79.96\%$ ($1101/1377$) & $95.64\%$ ($1317/1377$) & $272$ & $216/272$ ($79.41\%$) \\
\bottomrule
\end{tabular}%
}
\end{center}
\vspace{-0.8em}
\FloatBarrier

\section{Conclusion}

We introduced MILP-Evo, a framework for automatically designing MILP solver components as executable PySCIPOpt callbacks, which searches over native solver code and evaluates each candidate through end-to-end SCIP execution. This design keeps the discovered cut-selection and branching logic explicit, inspectable, and compatible with standard solver workflows. Our experiments show that LLM-guided executable search can discover competitive domain-specialized solver components across standard learn2branch families. The ablation and lineage analyses suggest that these gains come from jointly callback evolution, non-monotone reuse of useful program motifs, and contract-preserving repair. More broadly, MILP-Evo points to a practical path for automatic solver design, where the learned object is explicit algorithmic logic shaped by end-to-end solver behavior rather than a black-box policy.

\paragraph{Limitations.}
MILP-Evo currently targets two callback interfaces, cut selection and branching; although we explored node selection, primal heuristics, and separators, these extensions did not yield consistent runtime improvements under the present search setting. Transfer across instance scales is also uneven across families. Finally, LLM-guided program search remains sensitive to the model, prompts, sampling, budget, population state, and repair decisions; repair improves executability but does not guarantee semantic or solver-version robustness.

\clearpage
\bibliographystyle{plainnat}
\bibliography{ref}


\newpage
\appendix
\input{APP.tex}


\end{document}

%% file: APP.tex
\section{Implementation Details}

\subsection{Benchmark Instance Generation}

We generate the MILP instances using the same learn2branch-style generation protocol as \citet{gasse2019exact}. The released generation script writes CPLEX LP-format instances under problem-specific directories with \texttt{train}, \texttt{valid}, \texttt{transfer}, and \texttt{test} prefixes. For each benchmark family, MILP-Evo searches on Easy-scale training instances and evaluates the exported callback on held-out Easy, Medium, and Hard instance scales. The generator creates 1000 Easy-scale training instances and 200 Easy-scale validation instances per family; the evaluation scripts then select the held-out instances used for the reported transfer experiments.

The four benchmark families instantiate different MILP structures:
\begin{description}
\item[Set covering.]
Given a universe of elements and a collection of subsets whose union contains the universe, the set covering problem asks for a subcollection that still covers all elements. In our MILP instances, each binary variable indicates whether one subset is selected, and the objective is to obtain a cover with minimum total cost.

\item[Combinatorial auction.]
In a combinatorial auction, bidders submit bids on packages of discrete, heterogeneous items rather than only on individual items. The associated winner determination problem selects a compatible set of package bids that maximizes the auctioneer's revenue or total accepted value. In our MILP formulation, each binary variable corresponds to accepting one bid, and item-conflict constraints ensure that no item is assigned more than once.

\item[Capacitated facility location.]
Facility location problems ask where facilities should be placed or opened so that demand points can be served at low cost. In the capacitated variant used here, the model chooses which candidate facilities to open and how to route customer demand to open facilities. The objective combines fixed opening costs and transportation costs, while capacity constraints limit the total demand that each open facility may serve.

\item[Maximum independent set.]
In graph theory, an independent set is a set of vertices with no edge between any selected pair. The maximum independent set problem seeks an independent set of largest possible cardinality in a given graph. Our MILP uses one binary variable per vertex, edge constraints to prevent adjacent vertices from being selected together, and a cardinality-maximization objective.
\end{description}

Table~\ref{tab:instance-generation-hyperparams} reports the main scale parameters used for the Easy, Medium, and Hard instance families. Additional generator constants follow the learn2branch defaults used in our script: set-covering density is $0.05$ with objective coefficients sampled up to $100$,
the combinatorial-auction bundle-extension probability is $0.7$, the facility-location capacity-to-demand ratio is $5$, and independent-set graphs are generated by the Barabasi--Albert model with affinity $4$.

\begin{table}[h]
\centering
\small
\setlength{\tabcolsep}{8pt}
\renewcommand{\arraystretch}{1.18}
\caption{Instance generation hyperparameters for the benchmark families.}
\label{tab:instance-generation-hyperparams}
\begin{tabular}{p{0.34\linewidth}p{0.52\linewidth}}
\toprule
Benchmark & Hyperparameters \\
\midrule
Set covering
& \begin{tabular}[c]{@{}l@{}}
Easy: 500 rows, 1000 columns \\
Medium: 1000 rows, 1000 columns \\
Hard: 2000 rows, 1000 columns
\end{tabular} \\
\midrule
Combinatorial auction
& \begin{tabular}[c]{@{}l@{}}
Easy: 100 items, 500 bids \\
Medium: 200 items, 1000 bids \\
Hard: 300 items, 1500 bids
\end{tabular} \\
\midrule
Capacitated facility location
& \begin{tabular}[c]{@{}l@{}}
Easy: 100 facilities, 100 customers \\
Medium: 100 facilities, 200 customers \\
Hard: 100 facilities, 400 customers
\end{tabular} \\
\midrule
Maximum independent set
& \begin{tabular}[c]{@{}l@{}}
Easy: 500 nodes, affinity 4 \\
Medium: 1000 nodes, affinity 4 \\
Hard: 1500 nodes, affinity 4
\end{tabular} \\
\bottomrule
\end{tabular}
\end{table}

\subsection{Search and Evaluation Protocol}
\label{app:indset-workflow}

We use the independent-set benchmark as a representative example of the training and evaluation protocol. The same protocol is applied to the other problem families with problem-specific instance generators and callback evaluation files.

\paragraph{Training.}
For a given problem family, MILP-Evo searches over a population of executable Python artifacts. Each artifact jointly defines a cut selector and a branching rule, and both callbacks are registered in the same SCIP model during evaluation. The search is initialized with diverse LLM-generated candidates and then alternates between mutation, crossover, reflection, and repair. Candidate quality is never judged by the LLM itself: each program is compiled, checked against the callback contracts, executed inside SCIP on a small batch of training instances, and scored by mean process time when all instances are solved to optimality. Programs that violate the interface, raise callback errors, or fail the required solve condition receive infinite fitness.

The population is maintained as a quality-diversity archive. Language-model feedback is used to propose code edits and to describe candidates, while selection within the archive is determined by SCIP execution. This separation is important: the LLM explores the program space, but the solver supplies the fitness signal that drives selection.

To make the search interface concrete, Config~\ref{lst:indset-config} and Prompt~\ref{lst:indset-prompt} provide shortened excerpts from the independent-set configuration and task prompt. The configuration excerpt shows how the benchmark, search budget, and population structure are specified. The prompt excerpt shows how the LLM is instructed to produce one joint callback artifact and to respect the PySCIPOpt callback contracts. The full configuration files and prompts are included in the supplementary material.

\renewcommand{\lstlistingname}{Config}
\begin{lstlisting}[style=yamlcode,caption={Shortened independent-set search configuration excerpt.},label={lst:indset-config}]
mutation_rate: 0.4
crossover_rate: 0.6
batch_size: 10

instances:
  source: learn2branch
  problem: indset
  limit: 10
  scip_seed: 0

population:
  iterations: 400
  population_size: 20
  island_num: 4
  bins_num: 3
\end{lstlisting}

\renewcommand{\lstlistingname}{Prompt}
\begin{lstlisting}[style=appendixcode,caption={Shortened independent-set task prompt excerpt.},label={lst:indset-prompt}]
You are an expert in MILP solver design for maximum independent set
(stable set) instances. Your task is to co-evolve two SCIP plugins:
(1) a custom cut selector (CustomCutsel) that orders or filters
candidate cuts at separation, and
(2) a custom branching rule (CustomBranchingRule) that decides how to
branch on fractional LP candidates.

Both plugins run together on the same PySCIPOpt Model. One submission
must define both classes in the same module. CustomCutsel must return a
dictionary with keys cuts, nselectedcuts, and result, where cuts is only
a reordering or subset of the input cut pool. CustomBranchingRule must
branch when LP branching candidates exist and return SCIP_RESULT.BRANCHED.

Fitness is the mean process time to reach optimality over the evaluation
batch. Co-evolution means joint changes to cut selection and branching
must improve end-to-end SCIP behavior, not only one component in isolation.
\end{lstlisting}

\paragraph{Evaluation.}
After search, the best finite-fitness program for each family is exported as the family-specific policy. No further LLM calls or adaptation are performed during evaluation. The exported callbacks are loaded into fresh SCIP models and tested on held-out learn2branch transfer instances at Easy, Medium, and Hard scales. We use the same joint registration path as in training and report solver status, process time, and branch-and-bound node counts under the time limit described in the main text.

\section{Best Discovered Cut-Branch Callback}
\label{app:best-cut-branch}

This section presents the best joint cut-selection and branching callback artifact discovered by MILP-Evo. The artifact is an executable PySCIPOpt program registered inside SCIP during evaluation. It contains both the cut selector and the branching rule used for the corresponding benchmark family.
For compactness, we show the independent-set artifact in full because it is the setting where MILP-Evo obtains its strongest gains and where we also report detailed search dynamics in the main paper. The complete exported programs for all four benchmark families are included in the supplementary material under \texttt{problem/}.

\renewcommand{\lstlistingname}{Program}

\begin{lstlisting}[
	style=pythoncode,
	caption={Best discovered joint cut-selection and branching callback for independent set.},
	label={lst:best-indset}
]
from pyscipopt import Model, Branchrule, SCIP_RESULT
from pyscipopt.scip import Cutsel

class CustomCutsel(Cutsel):
    def __init__(self):
        super().__init__()

    def cutselselect(self, cuts, forcedcuts, root, maxnselectedcuts):
        if not cuts:
            return {"cuts": [], "nselectedcuts": 0, "result": SCIP_RESULT.SUCCESS}

        # Scoring: Blend efficacy with a tuned density penalty.
        # An exponent of 0.15 is used to balance Parent A's (0.4) sparsity 
        # and Parent B's (0.05) raw strength, aiming for fast LP solves.
        scored_cuts = []
        for c in cuts:
            eff = self.model.getCutEfficacy(c)
            nnz = c.getNNonz()
            # Density penalty (NNonz) keeps the LP sparse and iterations fast.
            score = eff / (nnz + 1.0)**0.15
            scored_cuts.append((c, score))

        # Reorder cuts by the hybrid score (descending)
        scored_cuts.sort(key=lambda x: x[1], reverse=True)
        
        # Gap-based selection: Filter cuts within 85% of the best score.
        # This was a successful strategy in both parents to maintain quality.
        best_score = scored_cuts[0][1]
        threshold = 0.85 * best_score
        
        n_pass = 0
        for _, score in scored_cuts:
            if score >= threshold:
                n_pass += 1
            else:
                break
        
        # Selection limit: Combine Parent B's 1/3 limit with Parent A's aggressiveness.
        # We cap selection to at most 30% of the pool or maxnselectedcuts.
        pool_limit = max(1, len(cuts) // 3)
        final_n = min(n_pass, maxnselectedcuts, pool_limit)
        
        return {
            "cuts": [x[0] for x in scored_cuts],
            "nselectedcuts": int(final_n),
            "result": SCIP_RESULT.SUCCESS
        }

class CustomBranchingRule(Branchrule):
    def __init__(self, model):
        super().__init__()
        self.model = model

    def branchexeclp(self, additional_cons):
        cands, sols, _, nlpcands, _, _ = self.model.getLPBranchCands()
        
        if nlpcands == 0:
            return {"result": SCIP_RESULT.DIDNOTRUN}

        # Hybrid Lexicographical Branching for Independent Set:
        # 1. getNLocksUp(): Corresponds to vertex degree in the conflict graph (Parent B's strength).
        # 2. getObj(): Prioritize high-weight nodes (critical for Maximum Weighted Independent Set).
        # 3. getIndex(): Use negative index to preserve deterministic tie-breaking (Parent A's strength).
        # 4. Fractionality: Standard tie-breaker to target most uncertain variables.
        
        best_cand = None
        best_score = (-1, -1.0, 1e12, -1.0) # (locks, obj, -index, frac)
        
        for i in range(nlpcands):
            v = cands[i]
            s = sols[i]
            
            # Metric Calculation
            locks = v.getNLocksUp()
            obj = v.getObj()
            idx = v.getIndex()
            frac = s * (1.0 - s) # maximized at 0.5
            
            current_score = (locks, obj, -idx, frac)
            
            if current_score > best_score:
                best_score = current_score
                best_cand = v
        
        if best_cand is not None:
            self.model.branchVar(best_cand)
            return {"result": SCIP_RESULT.BRANCHED}
        else:
            return {"result": SCIP_RESULT.DIDNOTRUN}
\end{lstlisting}

\section{Additional Experimental result}

\subsection{Result Variability and Stability}

Table~\ref{tab:main-results-std} compares standard deviations across methods on the same held-out instances used in the main evaluation. Because the test set is fixed for all methods, the reported deviations allow us to compare how much each method's runtime varies across the same group of instances. These deviations should not be interpreted as solver randomness alone: the instances within a benchmark family are fixed, but they can still differ substantially in difficulty. Consistent with the main table, the discussion below focuses on the primary comparison methods; GNN-GPU and COPT are included as external references.

For process time, MILP-Evo has the smallest standard deviation among the primary methods in half of the family-scale settings and is among the three lowest in most settings. The pattern is clearest on independent set and on the easy and medium combinatorial-auction instances, where low variance is paired with strong win counts and low mean runtime. This suggests that the improvement is not driven only by a few unusually easy instances: in these regimes, the discovered callback tends to reduce runtime consistently across the evaluation set. Similar runtime-stability behavior appears on the medium-scale set-covering and facility location instances.

The comparison also shows where this claim should be qualified. Very small runtime deviations for some baselines on hard instances can reflect consistent timeouts rather than robust solving, so standard deviation must be read together with mean time and win counts. Node counts require the same joint reading. Across several family-scale settings, methods that explore fewer branch-and-bound nodes are not necessarily faster, and MILP-Evo sometimes reaches a lower mean runtime while visiting more nodes. This pattern is consistent with a tradeoff between tree size and per-node overhead: a policy can reduce the number of nodes by spending more effort on branching or cut decisions, while a cheaper callback can traverse a larger tree but still finish sooner.

Thus, the larger node-count deviations of MILP-Evo in some settings should not be read as a direct contradiction of its runtime stability. They indicate that the evolved callbacks can produce more instance-dependent search trees, whereas the process-time columns show whether those trees are traversed efficiently enough to improve wall-clock performance. This distinction is most favorable on independent set and combinatorial auctions, where higher or more variable node counts often coincide with low mean runtime. On harder set-covering and facility-location instances, however, both runtime and node-count dispersion are larger, suggesting that transfer stability remains more problem-dependent outside the families where the learned solver behavior aligns most strongly with the instance structure.

\begin{table}[!t]
\centering
\tiny
\setlength{\tabcolsep}{2.0pt}
\renewcommand{\arraystretch}{0.82}
\caption{Main comparative results with standard deviations. Time and node counts are reported as mean$\pm$std over the same evaluation runs used for the main results. Wins and boldface follow the convention of Table~\ref{tab:main-results}: GNN-GPU and COPT are external references and are excluded from primary win-count and boldface comparisons.}
\label{tab:main-results-std}
\resizebox{\linewidth}{!}{%
\begin{tabular}{lccc@{\quad}ccc@{\quad}ccc}
\toprule
Indset: & \multicolumn{3}{c}{Easy} & \multicolumn{3}{c}{Medium} & \multicolumn{3}{c}{Hard} \\
\midrule
Model & Time(s) & Wins & Nodes & Time(s) & Wins & Nodes & Time(s) & Wins & Nodes \\
\cmidrule(lr){2-4}\cmidrule(lr){5-7}\cmidrule(lr){8-10}
FSB & $87.00{\pm}78.11$ & 1/50 & $30.2{\pm}27.5$ & $3318.97{\pm}789.79$ & 0/7 & $150.5{\pm}46.0$ & $3600.06{\pm}0.05$ & 0/0 & $53.9{\pm}16.5$ \\
\cmidrule(lr){2-4}\cmidrule(lr){5-7}\cmidrule(lr){8-10}
RPB & $9.49{\pm}3.77$ & 0/50 & $46.9{\pm}67.6$ & $194.74{\pm}189.02$ & 0/50 & $7442.9{\pm}8833.1$ & $2587.99{\pm}792.07$ & 0/15 & $75329.3{\pm}28439.3$ \\
GNN & $5.80{\pm}2.55$ & 0/50 & $62.2{\pm}79.1$ & $1673.24{\pm}1473.07$ & 0/36 & $19872.9{\pm}23832.4$ & $3600.01{\pm}0.01$ & 0/0 & $49010.6{\pm}9406.4$ \\
Hybrid & $6.47{\pm}1.45$ & 0/50 & $53.1{\pm}55.4$ & $1502.27{\pm}1143.77$ & 0/30 & $48887.3{\pm}37997.5$ & $3600.01{\pm}0.01$ & 0/0 & $42854.8{\pm}7286.9$ \\
GS4CO & $7.29{\pm}2.27$ & 0/50 & $54.1{\pm}50.7$ & $147.26{\pm}93.78$ & 0/50 & $2785.0{\pm}3313.0$ & $2057.08{\pm}955.49$ & 0/30 & $67083.8{\pm}42269.5$ \\
SCIP & $14.80{\pm}5.91$ & 0/50 & $164.6{\pm}141.2$ & $377.99{\pm}254.89$ & 0/50 & $4023.9{\pm}3369.8$ & $3155.52{\pm}715.13$ & 0/18 & $22613.1{\pm}6737.8$ \\
MILP-Evo & $\mathbf{2.57{\pm}1.04}$ & \textbf{49/50} & $84.7{\pm}84.0$ & $\mathbf{59.20{\pm}47.27}$ & \textbf{50/50} & $1051.9{\pm}1040.4$ & $\mathbf{806.17{\pm}679.78}$ & \textbf{50/50} & $17980.3{\pm}19471.1$ \\
\cmidrule(lr){2-4}\cmidrule(lr){5-7}\cmidrule(lr){8-10}
GNN-GPU & $4.63{\pm}1.24$ & -/50 & $62.2{\pm}79.1$ & $858.07{\pm}1214.30$ & -/45 & $57818.8{\pm}89060.7$ & $3600.00{\pm}0.00$ & -/0 & $107167.4{\pm}20945.0$ \\
COPT & $2.62{\pm}1.69$ & -/50 & $9.6{\pm}12.5$ & $62.18{\pm}49.60$ & -/50 & $707.7{\pm}724.9$ & $1531.27{\pm}1169.88$ & -/43 & $10279.2{\pm}7357.4$ \\
\midrule\midrule
Cauctions: & \multicolumn{3}{c}{Easy} & \multicolumn{3}{c}{Medium} & \multicolumn{3}{c}{Hard} \\
\midrule
Model & Time(s) & Wins & Nodes & Time(s) & Wins & Nodes & Time(s) & Wins & Nodes \\
\cmidrule(lr){2-4}\cmidrule(lr){5-7}\cmidrule(lr){8-10}
FSB & $10.80{\pm}4.39$ & 0/50 & $25.5{\pm}17.2$ & $286.99{\pm}220.88$ & 0/50 & $253.1{\pm}165.5$ & $3177.54{\pm}784.12$ & 0/14 & $912.3{\pm}231.9$ \\
\cmidrule(lr){2-4}\cmidrule(lr){5-7}\cmidrule(lr){8-10}
RPB & $3.70{\pm}1.33$ & 0/50 & $14.8{\pm}12.7$ & $21.69{\pm}8.98$ & 1/50 & $1169.0{\pm}1072.4$ & $\mathbf{271.44{\pm}171.11}$ & \textbf{36/50} & $17918.5{\pm}12358.2$ \\
GNN & $3.13{\pm}1.33$ & 0/50 & $105.8{\pm}65.0$ & $30.05{\pm}22.76$ & 0/50 & $1047.9{\pm}893.2$ & $662.39{\pm}451.27$ & 0/50 & $17571.3{\pm}12029.1$ \\
Hybrid & $2.07{\pm}0.65$ & 0/50 & $111.2{\pm}70.0$ & $21.26{\pm}15.91$ & 0/50 & $1210.7{\pm}1069.5$ & $718.47{\pm}565.48$ & 0/50 & $30404.5{\pm}24134.5$ \\
GS4CO & $3.07{\pm}1.46$ & 0/50 & $114.3{\pm}82.8$ & $42.55{\pm}41.19$ & 0/50 & $1353.1{\pm}1435.9$ & $1248.98{\pm}866.53$ & 0/45 & $24867.6{\pm}16419.1$ \\
SCIP & $3.57{\pm}1.34$ & 0/50 & $30.1{\pm}19.8$ & $44.96{\pm}24.76$ & 0/50 & $1169.9{\pm}984.0$ & $307.55{\pm}172.59$ & 3/50 & $15560.5{\pm}10578.3$ \\
MILP-Evo & $\mathbf{0.81{\pm}0.39}$ & \textbf{50/50} & $196.1{\pm}96.7$ & $\mathbf{11.79{\pm}8.53}$ & \textbf{49/50} & $1536.0{\pm}1087.2$ & $381.90{\pm}396.98$ & 11/50 & $34534.7{\pm}32040.0$ \\
\cmidrule(lr){2-4}\cmidrule(lr){5-7}\cmidrule(lr){8-10}
GNN-GPU & $1.60{\pm}0.43$ & -/50 & $105.8{\pm}65.0$ & $10.36{\pm}5.62$ & -/50 & $1047.9{\pm}893.2$ & $191.11{\pm}132.17$ & -/50 & $17571.7{\pm}12029.4$ \\
COPT & $2.18{\pm}0.74$ & -/50 & $126.8{\pm}118.9$ & $14.11{\pm}6.61$ & -/50 & $1542.2{\pm}1246.8$ & $161.34{\pm}97.05$ & -/50 & $19687.7{\pm}13324.3$ \\
\midrule\midrule
Setcover: & \multicolumn{3}{c}{Easy} & \multicolumn{3}{c}{Medium} & \multicolumn{3}{c}{Hard} \\
\midrule
Model & Time(s) & Wins & Nodes & Time(s) & Wins & Nodes & Time(s) & Wins & Nodes \\
\cmidrule(lr){2-4}\cmidrule(lr){5-7}\cmidrule(lr){8-10}
FSB & $29.07{\pm}32.68$ & 0/50 & $45.7{\pm}59.1$ & $1291.03{\pm}1153.67$ & 0/45 & $929.5{\pm}640.3$ & $3600.03{\pm}0.02$ & 0/0 & $658.3{\pm}330.6$ \\
\cmidrule(lr){2-4}\cmidrule(lr){5-7}\cmidrule(lr){8-10}
RPB & $11.20{\pm}4.61$ & 0/50 & $129.6{\pm}249.7$ & $\mathbf{162.69{\pm}233.22}$ & \textbf{29/50} & $10349.9{\pm}17486.7$ & $2401.92{\pm}876.57$ & \textbf{19/20} & $118275.4{\pm}52191.2$ \\
GNN & $11.32{\pm}8.39$ & 3/50 & $172.7{\pm}182.5$ & $487.64{\pm}709.82$ & 0/50 & $5990.5{\pm}8593.8$ & $3500.75{\pm}300.99$ & 0/5 & $21036.2{\pm}3297.5$ \\
Hybrid & $\mathbf{8.26{\pm}3.59}$ & \textbf{29/50} & $201.9{\pm}210.3$ & $171.41{\pm}225.19$ & 15/50 & $7240.7{\pm}9825.5$ & $\mathbf{2359.13{\pm}702.19}$ & 0/10 & $63408.9{\pm}28316.7$ \\
GS4CO & $13.89{\pm}10.38$ & 0/50 & $249.1{\pm}268.4$ & $609.94{\pm}836.20$ & 0/45 & $10303.9{\pm}13451.6$ & $2937.46{\pm}189.53$ & 0/5 & $30500.2{\pm}5994.1$ \\
SCIP & $14.18{\pm}8.24$ & 0/50 & $124.5{\pm}229.6$ & $274.02{\pm}309.37$ & 0/50 & $8361.4{\pm}12497.3$ & $2955.81{\pm}982.01$ & 0/20 & $115143.9{\pm}50744.5$ \\
MILP-Evo & $9.03{\pm}5.47$ & 18/50 & $232.6{\pm}227.9$ & $166.69{\pm}176.54$ & 6/50 & $10109.2{\pm}14585.5$ & $2895.95{\pm}1040.37$ & 5/24 & $133883.2{\pm}73442.1$ \\
\cmidrule(lr){2-4}\cmidrule(lr){5-7}\cmidrule(lr){8-10}
GNN-GPU & $5.31{\pm}2.14$ & -/50 & $172.7{\pm}182.5$ & $102.20{\pm}138.75$ & -/50 & $5990.5{\pm}8593.8$ & $2492.68{\pm}1269.69$ & -/20 & $102635.8{\pm}54080.0$ \\
COPT & $6.88{\pm}3.32$ & -/50 & $297.4{\pm}412.9$ & $99.64{\pm}115.85$ & -/50 & $12207.1{\pm}17679.6$ & $2718.71{\pm}1135.03$ & -/25 & $175771.1{\pm}80573.4$ \\
\midrule\midrule
Facilities: & \multicolumn{3}{c}{Easy} & \multicolumn{3}{c}{Medium} & \multicolumn{3}{c}{Hard} \\
\midrule
Model & Time(s) & Wins & Nodes & Time(s) & Wins & Nodes & Time(s) & Wins & Nodes \\
\cmidrule(lr){2-4}\cmidrule(lr){5-7}\cmidrule(lr){8-10}
FSB & $81.02{\pm}59.56$ & 0/50 & $71.0{\pm}61.1$ & $806.48{\pm}669.60$ & 0/50 & $233.9{\pm}223.3$ & $2108.21{\pm}1143.00$ & 0/41 & $109.6{\pm}77.2$ \\
\cmidrule(lr){2-4}\cmidrule(lr){5-7}\cmidrule(lr){8-10}
RPB & $36.03{\pm}25.70$ & 16/50 & $90.4{\pm}110.2$ & $305.37{\pm}232.06$ & 6/50 & $588.6{\pm}645.0$ & $\mathbf{607.57{\pm}221.14}$ & 15/50 & $260.1{\pm}171.7$ \\
GNN & $50.48{\pm}41.74$ & 0/50 & $287.6{\pm}271.4$ & $478.05{\pm}510.82$ & 7/50 & $881.4{\pm}827.2$ & $669.15{\pm}321.01$ & 8/50 & $462.3{\pm}300.8$ \\
Hybrid & $\mathbf{35.09{\pm}28.81}$ & \textbf{17/50} & $253.2{\pm}244.9$ & $325.92{\pm}284.35$ & 14/50 & $895.3{\pm}896.8$ & $631.09{\pm}296.42$ & \textbf{17/50} & $456.6{\pm}297.0$ \\
GS4CO & $68.69{\pm}64.17$ & 0/50 & $273.6{\pm}267.3$ & $558.56{\pm}475.23$ & 1/50 & $933.8{\pm}861.2$ & $731.28{\pm}387.37$ & 4/50 & $449.9{\pm}294.0$ \\
SCIP & $38.27{\pm}22.93$ & 5/50 & $99.5{\pm}84.3$ & $305.10{\pm}241.99$ & 2/50 & $541.1{\pm}606.3$ & $675.33{\pm}284.41$ & 3/50 & $194.6{\pm}148.3$ \\
MILP-Evo & $35.66{\pm}23.35$ & 12/50 & $235.5{\pm}237.6$ & $\mathbf{266.21{\pm}218.47}$ & \textbf{20/50} & $926.8{\pm}957.4$ & $761.88{\pm}360.30$ & 3/50 & $429.1{\pm}288.0$ \\
\cmidrule(lr){2-4}\cmidrule(lr){5-7}\cmidrule(lr){8-10}
GNN-GPU & $33.44{\pm}24.30$ & -/50 & $287.6{\pm}271.4$ & $267.77{\pm}220.73$ & -/50 & $881.4{\pm}827.2$ & $546.13{\pm}258.44$ & -/50 & $462.3{\pm}300.8$ \\
COPT & $8.20{\pm}5.79$ & -/50 & $221.0{\pm}194.4$ & $31.56{\pm}27.88$ & -/50 & $572.0{\pm}587.5$ & $58.06{\pm}27.46$ & -/50 & $262.8{\pm}164.6$ \\
\bottomrule
\end{tabular}%
}
\end{table}
\FloatBarrier

\subsection{Additional Search Curves}

\begin{figure}[!t]
\centering
\begin{minipage}{0.48\linewidth}
\centering
\includegraphics[width=\linewidth]{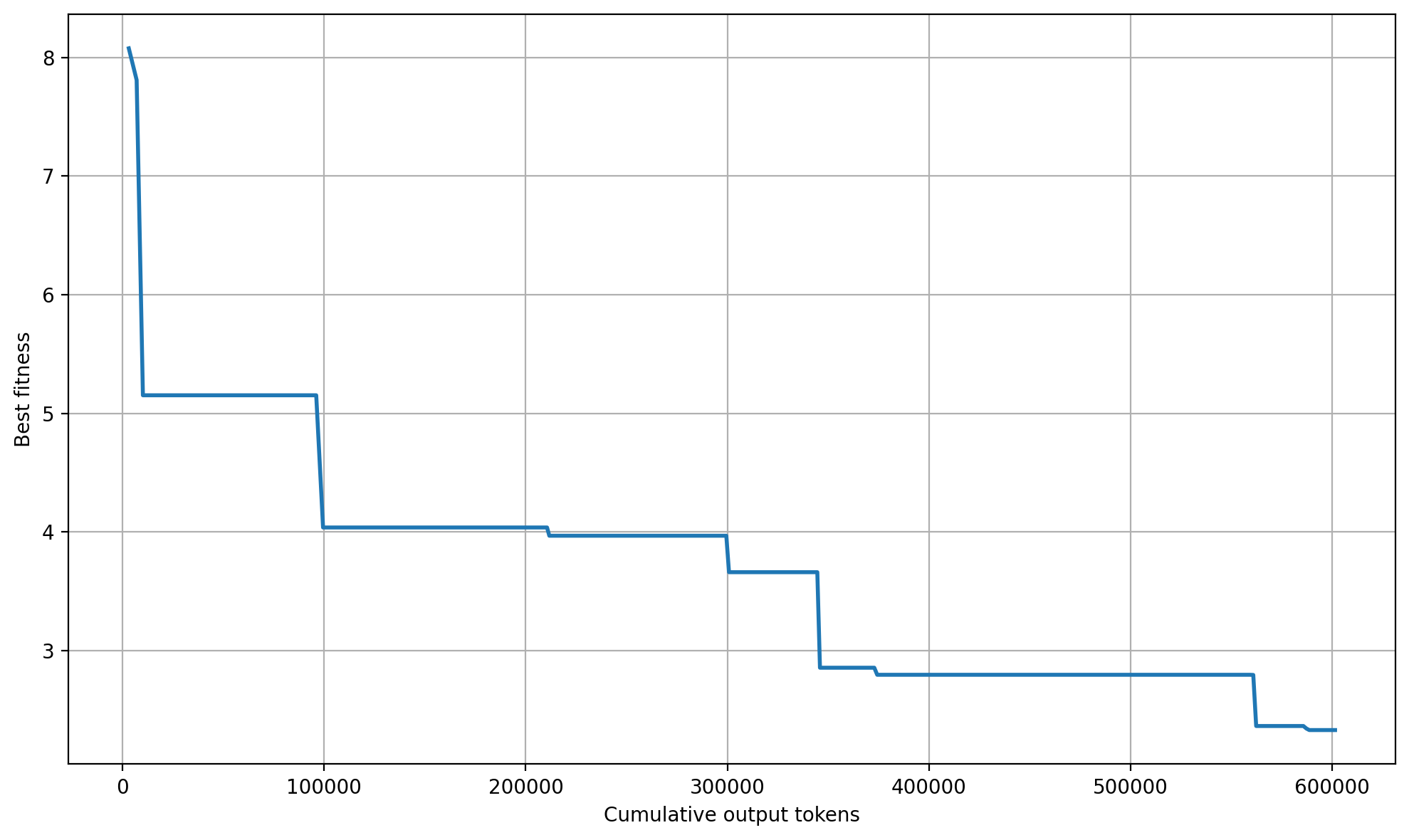}\\[-0.3em]
{\footnotesize (a) Maximum independent set}
\end{minipage}\hfill
\begin{minipage}{0.48\linewidth}
\centering
\includegraphics[width=\linewidth]{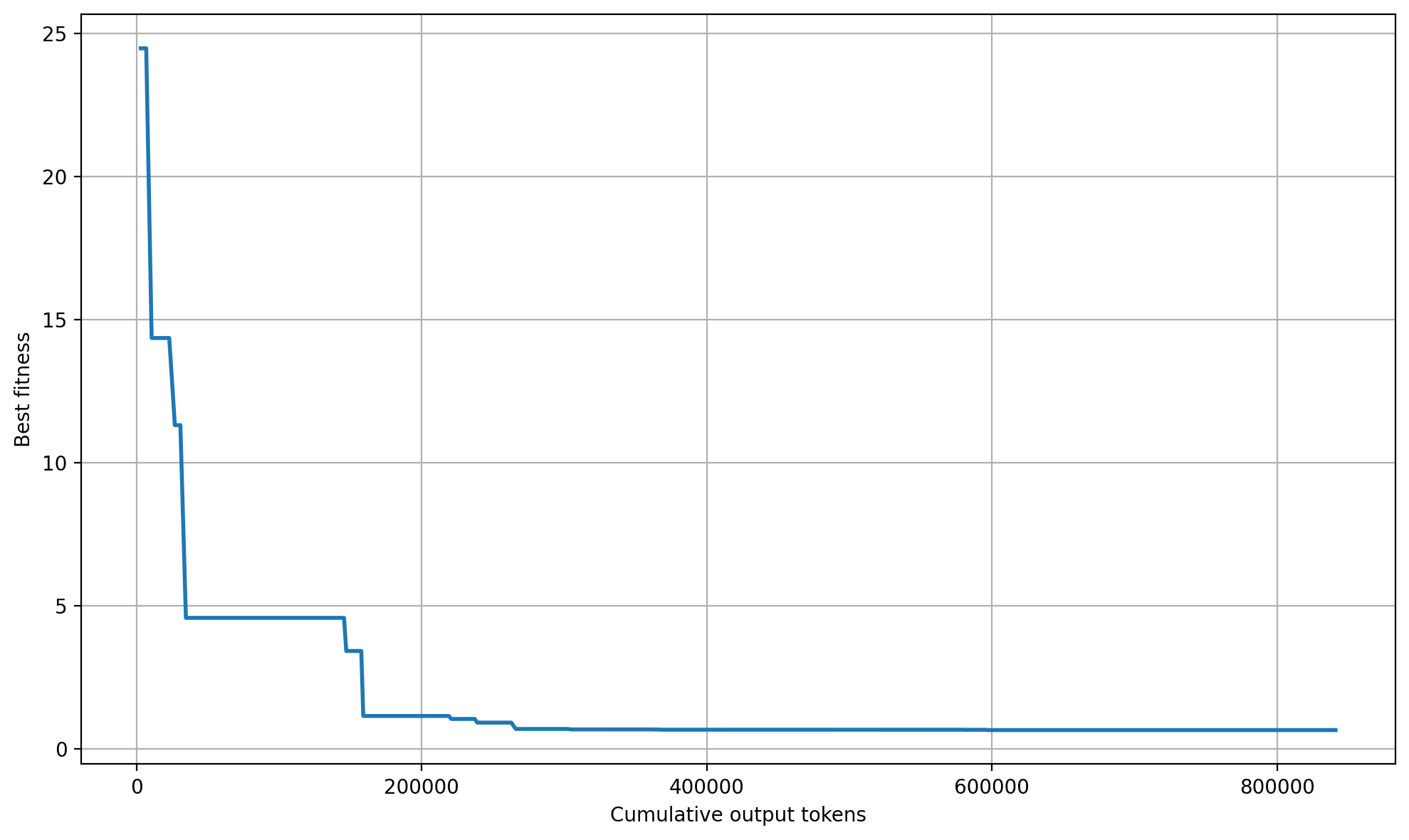}\\[-0.3em]
{\footnotesize (b) Combinatorial auction}
\end{minipage}

\vspace{0.6em}
\begin{minipage}{0.48\linewidth}
\centering
\includegraphics[width=\linewidth]{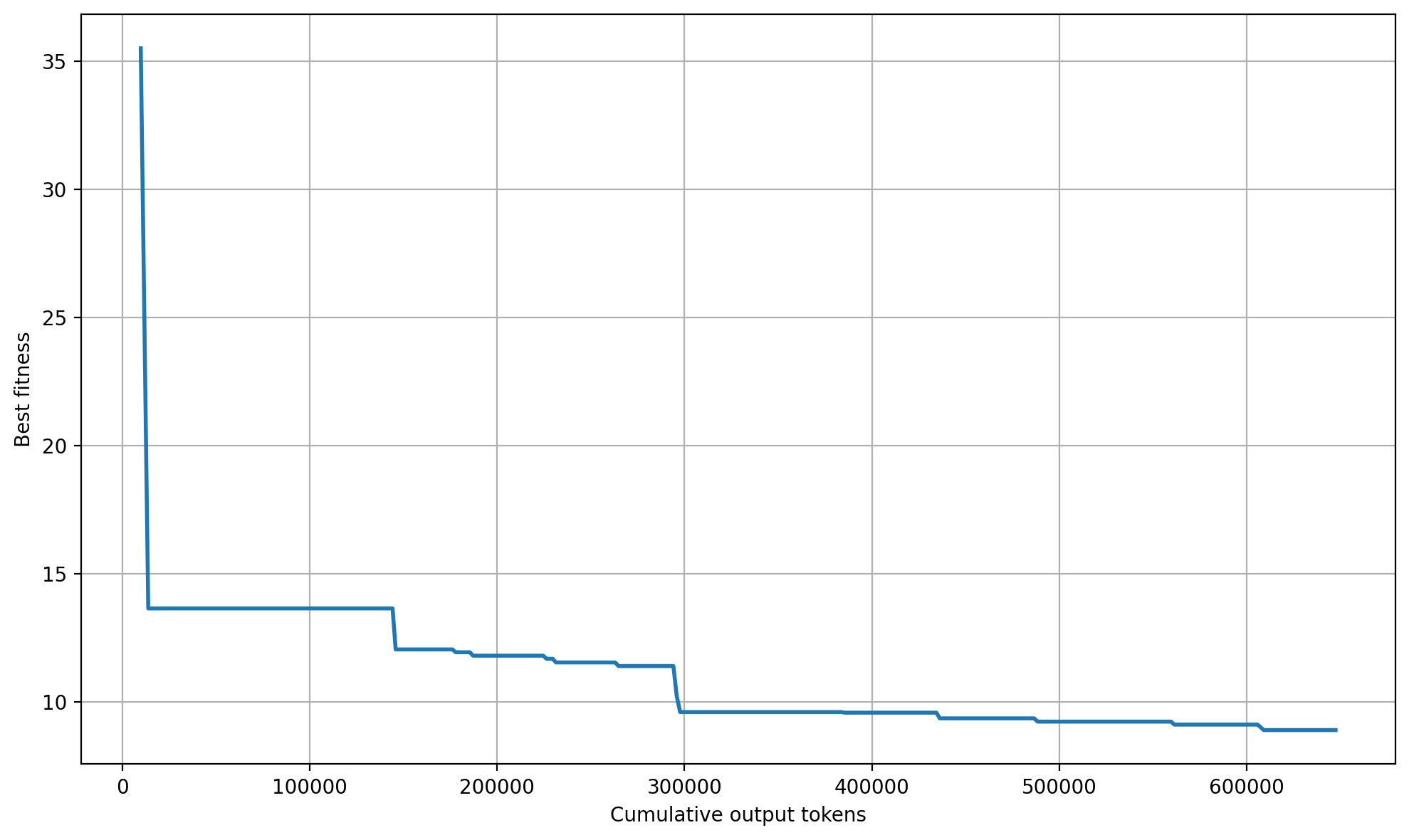}\\[-0.3em]
{\footnotesize (c) Set covering}
\end{minipage}\hfill
\begin{minipage}{0.48\linewidth}
\centering
\includegraphics[width=\linewidth]{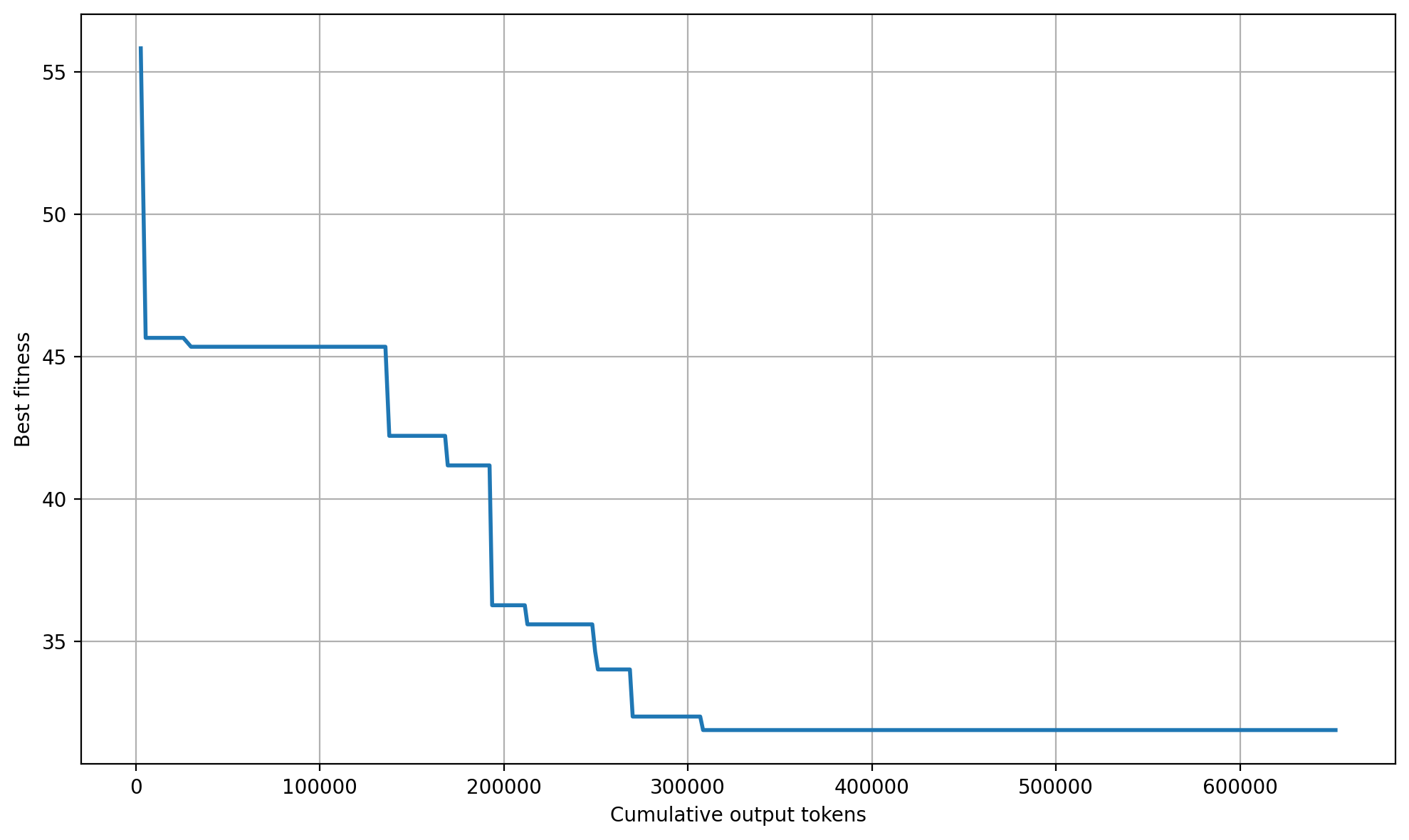}\\[-0.3em]
{\footnotesize (d) Capacitated facility location}
\end{minipage}
\caption{Additional search curves across the four benchmark families. Lower
fitness values indicate better candidate callback programs under the search
objective.}
\label{fig:additional-search-curves}
\end{figure}

Figure~\ref{fig:additional-search-curves} reports the search traces for all four benchmark families. Each curve tracks the best fitness found during the evolutionary search, providing a family-level view of how useful callback programs emerge beyond the independent-set example shown in the main text. Since the plotted value is the best archived fitness, the curves are monotone and should be read as discovery traces rather than as the performance of every sampled program.

The curves share a staircase pattern: long plateaus are interrupted by a small number of large drops. This behavior is consistent with program search in a constrained solver API. Many candidate edits are either invalid, behaviorally neutral, or only locally useful; occasionally, however, a structural change in cut filtering, branching priority, or their interaction produces a qualitatively better callback. After such a candidate is found, the archive retains it and subsequent search starts from a stronger incumbent.

The family-level dynamics differ in ways that match the quantitative results. Combinatorial auctions improve rapidly in the early part of search and then enter a long low-fitness plateau, suggesting that the search finds an effective solver strategy relatively quickly and later iterations mostly refine it. Independent set also improves early, but unlike combinatorial auctions it continues to obtain meaningful late-stage gains; this is consistent with the strong transfer performance of the final independent-set callback. Set covering and facility location show longer plateaus and smaller late improvements. Their curves indicate that the search still discovers better programs, but improvements arrive less frequently and the final incumbents are less dominant, matching the more mixed results in Table~\ref{tab:main-results-std}.

Overall, the search curves indicate that progress is uneven and problem-dependent. All four families exhibit incumbent improvements after initialization, but the timing and magnitude of these improvements vary substantially. Combinatorial auctions and independent set show clearer early or sustained gains, whereas set covering and facility location spend longer periods on plateaus and obtain smaller late-stage improvements. This pattern is consistent with the evaluation results, where transfer performance is strongest on the families with clearer search progress.

\FloatBarrier

\section{Broader Impacts}

MILP-Evo studies automatic design of explicit solver components for mixed-integer linear programming. The most direct positive impact is computational: better domain-specialized cut-selection and branching rules can reduce the time and energy required to solve recurring optimization problems. Because the discovered artifacts are executable PySCIPOpt callbacks rather than opaque neural predictors, they can be inspected, modified, and deployed within standard solver workflows. This transparency may make learned solver customization more usable for researchers and practitioners who need to audit optimization logic before using it in production systems.

The same capability also has potential risks. Faster MILP solving can improve decision systems in domains such as logistics, scheduling, manufacturing, network design, and resource allocation; some of these deployments may affect people if the underlying mathematical model encodes biased objectives, incomplete constraints, or sensitive business rules. Our method optimizes solver behavior, not the social desirability of the optimization model being solved. Thus, an automatically designed solver component should not be interpreted as a guarantee that the resulting decision process is fair, safe, or appropriate for a high-stakes application.

There are also technical risks specific to LLM-generated solver code. A candidate callback may be syntactically valid and fast on benchmark instances while still containing brittle assumptions, solver-version dependencies, or edge-case behavior that appears only on unseen instance distributions. We mitigate these risks in the research setting by using public synthetic benchmark families, running candidates inside SCIP with explicit callback-contract checks, assigning invalid or unsuccessful runs infinite fitness during search, and evaluating the final artifact by end-to-end solver execution. For any deployment beyond benchmark evaluation, we recommend manual inspection of the exported callback, stress testing on representative and adversarial instances, solver-version pinning, and domain-specific review of the MILP formulation and objective.

This work does not use private data, human-subject data, or scraped personal information. The experiments are conducted on standard synthetic MILP benchmark families, and the proposed method does not release a model that directly generates natural-language content for end users.

\section{Assets and Licenses}

Table~\ref{tab:assets-licenses} lists the existing assets used in this work and
their licenses or terms. We retain third-party notices and do not redistribute
commercial solvers or code whose redistribution terms are not explicit.

\begin{table}[htbp]
\centering
\small
\setlength{\tabcolsep}{5pt}
\renewcommand{\arraystretch}{1.05}
\caption{Used assets and their licenses.}
\label{tab:assets-licenses}
\resizebox{\linewidth}{!}{%
\begin{tabular}{llll}
\toprule
Type & Asset & License or terms & Usage \\
\midrule
Dataset/Code & learn2branch~\citep{gasse2019exact} & MIT License & Instance generation; evaluation \\
Code & Hybrid-learn2branch~\citep{gupta2020hybrid} & MIT License & Baseline evaluation \\
Code & GS4CO~\citep{kuang2024rethinking,kuang2024towards} & No explicit root license & Baseline evaluation \\
Code & PySCIPOpt 6.1.0~\citep{maher2016pyscipopt} & MIT License & Callback implementation \\
Solver & SCIP 10.0.2~\citep{achterberg2009scip} & Apache-2.0 License & MILP solving \\
Solver & SoPlex 8.0.2 & Apache-2.0 License & LP solving \\
Solver & COPT~\citep{copt} & Proprietary commercial solver; used under a COPT academic license & External reference \\
\bottomrule
\end{tabular}
}
\end{table}

All benchmark instances used in this work are synthetic MILP instances from
standard learn2branch-style distributions. They do not contain personal data,
human-subject data, copyrighted media, or scraped web content.